\documentclass[letterpaper]{article} 

\usepackage{aaai24}  
\usepackage{times}  
\usepackage{helvet}  
\usepackage{courier}  
\usepackage[hyphens]{url}  
\usepackage{graphicx} 
\usepackage{natbib}  
\usepackage{caption} 
\usepackage{algorithm}
\usepackage{algorithmic}

\usepackage{newfloat}
\usepackage{listings}
\DeclareCaptionStyle{ruled}{labelfont=normalfont,labelsep=colon,strut=off} 
\floatstyle{ruled}
\newfloat{listing}{tb}{lst}{}
\floatname{listing}{Listing}
\title{
The Inhibitor: ReLU and Addition-Based Attention for Efficient Transformers 
\\ under Fully Homomorphic Encryption on the Torus
}

\author {
    Rickard Brännvall\textsuperscript{\rm * $\dagger$ } and
    Andrei Stoian\textsuperscript{\rm $\ddagger$}
}
\affiliations {
    \textsuperscript{\rm $\dagger$}Computer Science Department, RISE Research Institutes of Sweden\\
    \textsuperscript{\rm $\ddagger$}Zama, Machine Learning Group, 
    Paris, France\\ 
    rickard.brannvall@ri.se, 
    andrei.stoian@zama.ai \\
    (Presented at FHE.org Conference 2024)
    }

\usepackage{booktabs}       
\usepackage{amsfonts}       
\usepackage{xcolor}         
\usepackage{csvsimple}
\usepackage{amsmath,amssymb}

\newcommand{\code}{\texttt}

\begin{document}

\maketitle

\begin{abstract}
To enhance the computational efficiency of quantized Transformers, we replace the dot-product and Softmax-based attention with an alternative mechanism involving addition and ReLU activation only. This side-steps the expansion to double precision often required by matrix multiplication and avoids costly Softmax evaluations but maintains much of the core functionality of conventional dot-product attention. It can enable more efficient execution and support larger quantized Transformer models on resource-constrained hardware or alternative arithmetic systems like homomorphic encryption. 
Training experiments on four common benchmark tasks show test set prediction scores comparable to those of conventional Transformers with dot-product attention. Our scaling experiments also suggest significant computational savings, both in plaintext and under encryption. 
In particular, we believe that the ReLU and addition-based attention mechanism examined in this paper may enable privacy-preserving AI applications operating under homomorphic encryption by avoiding the costly multiplication of encrypted variables.
\end{abstract}

\section{Introduction}

Transformer models \cite{Vasvani2017} have achieved great success in various machine-learning tasks and are the foundation for modern large-scale pre-trained Language Models such as BERT \cite{BERT},  
RoBERTa \cite{RoBERTa}, XLNet \cite{XLNet}, Transformer-XL \cite{Transformer-XL}, and the GPT family \cite{gpt2019,gpt2020}. 
These models have enabled leaps in performance on many NLP tasks, as for example experienced by the millions that use the ChatGPT application \cite{chatgpt}. 
Similarly, the Vision-Transformer (ViT) \cite{ViT2020}, and similar models outperform CNNs on many computer vision tasks such as image classification \cite{swin2021} and object detection \cite{carion2020}.

The impressive gains in performance, however, come at a price. State-of-the-art Transformers are extremely large, having parameters in the hundreds of billions. 
The requirements for memory and computational power have become prohibitive, not only for deployment on resource-constrained embedded systems, but sometimes even in data centers, and their considerable energy usage is widely discussed in society. 
Neural network quantization \cite{Zhou2016,Jacob_2018,Krishnamoorthi2018} is a method to address this problem, as it reduces memory usage by employing lower-bit precision for weight and activation tensors. This not only decreases inference time but also increases energy efficiency through the adoption of low-bit fixed-point arithmetics instead of floating-point operations \cite{Horowitz_2014}.
Other approaches to improve Transformer's scalability and environmental sustainability include 
Pruning \cite{Han2015,Han2016}, 
Knowledge Distillation \cite{Hinton2015,DistilBERT} and Neural Architecture Search \cite{Chen_2021}. 
This work discusses changes to the architecture's fundamental building blocks towards the same end. 

Transformers use an attention mechanism to identify connections between different elements in a sequence, which conventionally takes the dot-product between the \textit{query} and \textit{key} matrices before passing the results through a Softmax function. 
Many alternative formulations have been proposed, such as the Fastformer \cite{wu2021fastformer} based on additive attention, or the ReLUformer \cite{ReLUformer} that uses ReLU activation in place of Softmax.
These architectures share elements with the alternative formulation that is discussed in this work, however, they still apply attention by a matrix multiplication with the \textit{value} matrix. 

Extending our work with ReLU and addition-based gated RNNs \cite{brannvall2023relu}, we proposed the Inhibitor attention mechanism \cite{brannvall2024inhibitor_studentabstract}, which is designed to not rely on variable-to-variable multiplication and Softmax activation. 
These operations are more costly than constant-to-variable multiplications (also known as literal multiplication) and ReLU on many computing hardware.
Even more so, they are particularly challenging operations under Fully Homomorphic Encryption (FHE), an arithmetic system that permits calculation with encrypted variables without access to the secret key.
This article includes our previous results and gives more detail on the FHE implementation.

\section{Preliminaries}

TFHE, or Fully Homomorphic Encryption on the Torus \cite{Chillotti_2019}, is distinguished from other FHE schemes because it proposes the so-called Programmable Bootstrapping (PBS) mechanism. 
The PBS does a blind rotation over a predefined table of all the permittable values in a message space, e.g $[-N, N-1]$, landing on the coefficient whose index is closest to the original cipher text $x$. By filling the lookup table with values of a function evaluated over the original message space, $f(x)$, one instead obtains the PBS, which both reduces noise and evaluates a univariate function on a discrete set of points. 
It is also relatively fast compared to alternative FHE schemes.

TFHE does not natively have multiplication between ciphertexts, but it can be constructed by the application of two PBS operations
\begin{equation}
    a\, b = \mathrm{PBS}\left(f; a+b\right)
         - \mathrm{PBS}\left(f; a-b\right)
    \label{eq:mult_by_PBS_construction}
\end{equation}
where by $\mathrm{PBS}\left(f; x\right)$, we mean applying a table that corresponds to the function $f$ with argument $x$, which is 
\begin{equation}
    f(x) = \frac{x^2}{4}
    \label{eq:mult_by_PBS_function}
\end{equation}
in this case of multiplication by PBS. 

In order for the FHE scheme to be able to evaluate a circuit correctly, all the possible values that can occur for each component must be inside the allotted message space. 
The TFHE scheme operates on several different cipher-texts, namely LWE \cite{Regev2009}, RLWE \cite{Lyubashevsky2013_RLWE}, GLWE \cite{Brakerski_2012}, and generalizations of GSW \cite{Gentry2013_GSW}, each governed by specific parameters. For LWE ciphertexts, a dimension parameter is required, while GLWE ciphertexts rely on both polynomial size and dimension parameters. 

A framework for parameter optimization in TFHE proposed by \cite{Bergerat2023_ParamOpt} distinguishes such macro-parameters from the micro-parameters used exclusively within FHE operators, such as the decomposition base and the number of decomposition levels required for a PBS. To optimize FHE operators, both noise and cost models come into play. The noise model simulates noise evolution across an operator, while the cost model serves as a metric to minimize encompassing factors like execution time, power consumption, or price. As the selection of micro and macro parameters significantly impacts cost and noise growth for each operation, they demand prudent and deliberate choices. 
For all numerical experiments in this paper, we use parameters automatically selected by the Concrete Python Compiler \cite{Concrete_Python_Compiler} based on the TFHE parameter optimization framework by \cite{Bergerat2023_ParamOpt}. 

\paragraph{Privacy Preserving Machine Learning.}
Despite the computational challenges, variants of Fully Homomorphic Encryption have been successfully applied to achieve Privacy Preserving Machine Learning (PPML). 
Such efforts include hyperplane decision-based classification, Naive Bayes classification, decision tree classification \cite{Bost2015MachineLC, EmTopicsComp8260844}, 
classification and regression using Random Forests as demonstrated in Cryptotrees \cite{huynh2020cryptotree}, 
and more recent developments of efficient nearest neighbor classifiers \cite{Chakraborty_2022}.

Early approaches for Neural Networks like Cryptonets \cite{CryptoNets2016} focused on achieving blind, noninteractive classification by applying a leveled homomorphic encryption scheme \cite{Brakerski_2012} to the network inputs and used the square function as an activation function replacement to accommodate the limitations of the underlying encryption scheme. 
Subsequent works \cite{Zhang_2016} adopted higher-degree polynomials that approximate conventional activation functions like sigmoid and tanh on a range.
However, leveled homomorphic approaches have limited scalability, and as the number of layers in a neural network increases, the overall performance of homomorphic classification becomes prohibitive. 

To overcome the limitations of previous approaches, a hybrid cryptographic scheme \cite{Juvekar2018} was proposed that combined homomorphic encryption (HE) for processing linear layers with garbled circuits (GC) for computing activations in convolutional neural networks (CNNs). 
The garbled circuit is a cryptographic protocol that enables secure computation in which two  parties can jointly evaluate a function over their private inputs \cite{Yao_1982}.
This hybrid approach suffered from long inference latency due to slow GC-based activations, especially for tanh and sigmoid that required large circuits.

The emerging TFHE scheme \cite{Chillotti2016} showed promise for noninteractive PPML because of its fast bootstrap procedure and capacity to evaluate look-up tables. 
Based on this scheme, FHE–DiNN, a novel framework for homomorphic evaluation of discretized neural networks, was proposed \cite{Bourse2018_FHEofDDNN}. It featured linear complexity in the network depth and overcomes the scale limitations of leveled approaches, and applies the sign function for neuron activation by using bootstrapping. Empirical results demonstrate accurate classification by fully connected neural networks of encrypted MNIST dataset images with over 96\% accuracy in under 1.7 seconds.
Also using TFHE, a shift-accumulation-based convolutional deep neural network (SHE) \cite{SHE2019} was proposed to overcome the limitations of using polynomial approximations by implementing ReLU activations and max pooling. It employed logarithmic quantization to replace expensive FHE multiplications with cheaper shifts and claim state-of-the-art inference accuracy and reduced inference latency on MNIST and CIFAR-10 image classification data sets.

\paragraph{Computational Efficiency.}
Different operations within neural networks have distinct power and energy requirements, which can vary based on the underlying computer architecture \cite{Parhami_book_2010}. 
Addition is a relatively straightforward operation that can be executed in a single instruction. 
It is considered relatively inexpensive also under FHE.

Literal constants are fixed values that can be encoded in program instructions, while variables are undetermined at compile time. Literal multiplication can be optimized by the compiler using techniques like precomputed values or bit-shifting, resulting in faster execution and lower cost compared to variable multiplication also in clear text. Variable multiplication entails additional memory access to retrieve values, which can significantly impact energy consumption, as RAM access can be several orders of magnitude more energy-intensive than computation \cite{Horowitz_2014}.
While multiplication by literals is natively supported in TFHE, the multiplication of two encrypted variables is constructed using two Programmable Bootstrap (PBS) operations, which is much more expensive and can dominate the overall computational cost.

Activation functions play a vital role in neural networks. Functions like sigmoid require complex mathematical operations even in clear text, including exponentials and divisions, which are computationally expensive. On the other hand, the ReLU (Rectified Linear Unit) activation function has simpler implementations involving threshold comparisons, enabling more efficient execution on various hardware architectures.
Univariate non-linear activation functions can be evaluated in TFHE through the Programmable Bootstrap (PBS) operation, although it incurs significant computational costs. 

As the Concrete Python library for TFHE only supports integers natively, quantization becomes necessary when implementing neural network architectures: inputs, weights, and activations all need to be projected from floating-point values onto the integers. Furthermore, since any deep neural network with non-trivial activation functions will require PBS, the maximum precision of the TFHE table look-up implementation will be a limiting factor. At the time of the experiments, this was $2^7=128$ different values, i.e.,\ integer 7-bit precision\footnote{Later versions support up to 16-bit precision.}. Precision also directly impacts execution time as larger table lookups imply slower inference.

\section{Method}

\textbf{The Transformer} architecture consists of repeated blocks of self-attention and feed-forward networks (FFNs). In the simple decoder application, each block takes as input a sequence embedding, $X\in \mathbb{R}^{n\times d}$, where $n$ is the length and $d$ is the dimension. The embedding is transformed into \textit{query}, \textit{key} and \textit{value}, by multiplication with weight matrices according to $Q=X W_Q$, $K=X W_K$, and $V=X W_V$. 
\textit{Attention scores} are calculated by passing the scaled dot-product of \textit{query} and \textit{key} through the Softmax function, 
\begin{equation}
    S = \mathrm{Softmax}\left(Q K^T / \sqrt{d} \right)
    \label{eq:dotprod}
\end{equation}
which is then used to form a weighted sum of \textit{values} as output, $H=S\,V$, exploiting that each row of $S$ is normalized to sum to one. The FFN has two fully connected layers with weight matrix multiplication and ReLU activation,
\begin{equation}
    H = (X W_1^T + b_1)^+ W_2 + b_2
\end{equation}
where $W_1,W_2$ are weight matrices and $b_1,b_2$ are bias vectors. We use the short-hand notation $x^+=\max(0,x)$ for the ReLU function. 
Each Transformer block typically also has one or more layer normalizations \cite{layernorm} that stabilize the gradient flow and contribute to regularizing the training. 

\textbf{The Inhibitor} alternatively proposes to calculate \textit{attention scores} as
\begin{equation}
    Z_{ij} = \sum_{k} \frac{1 }{\gamma} 
        \left|Q_{ik} - K_{jk}\right|
    \label{eq:manhattan}
\end{equation}
where mixing between \textit{query} and \textit{key} is based on their absolute difference instead of dot-product. We have thus replaced the cosine (dot-prod) distance of conventional attention with the Manhattan distance. In place of Softmax multiplication, we subtract \textit{attention scores} from \textit{values} inside a ReLu function, 
\begin{equation}
    H_{ik} = \sum_j \left( V_{jk} - Z_{ij}  \right)^+
    \label{eq:inhibition}
\end{equation}
such that entries of $V$ for which $Z$ are large are zeroed out and do not contribute to the sum. We call this mechanism \textit{inhibition} as it is reminiscent of subtractive inhibition in biological neurons. FFN and normalization are left unchanged.

\textit{Shifted inhibition score.} The Manhattan score of equation \ref{eq:manhattan} can only take the value zero for rows where $Q$ and $V$ are identical. To make it easier to obtain a zero inhibition score, which allows entries of the \textit{value} matrix, $V$, to pass unmodified through equation \ref{eq:inhibition} we test using a slightly modified inhibition score, $Z'= (Z-\alpha)^+$, which applies a constant shift $\alpha \ge 0$ to the Manhattan score.

\begin{table}
\centering
\small
\begin{filecontents*}{all_results.txt}
Benchmark Tasks,Adding,MNIST,IMDB,IAMW
Dot-prod Attention,0.11\%,98.2\%,87.2\%,17.9
Inhibitor Attention,0.12\%,97.9\%,87.3\%,18.1
\end{filecontents*}
\csvautobooktabular{all_results.txt}
\caption{
The Inhibitor shows comparable performance to conventional attention for Transformers trained on four standard tasks (for mse, acc, acc, and edit distance, respectively).
}
\label{table:results_summary}
\end{table}

\section{Results}

\begin{table*}
\centering
\small
\begin{filecontents*}{stats_beta4_2_0_2_2.txt}
TFHE Parameters,T,lweDim,baseLog,level,polySize,int,uint
Inhibitor Attention,2,816,23,1,2048,5,4
Dot-prod Attention,2,817,23,1,2048,6,7
Inhibitor Attention,4,875,22,1,4096,6,5
Dot-prod Attention,4,834,23,1,2048,7,7
Inhibitor Attention,8,795,22,1,4096,5,5
Dot-prod Attention,8,792,22,1,4096,7,8
Inhibitor Attention,16,883,22,1,4096,6,6
Dot-prod Attention,16,794,15,2,4096,8,8
\end{filecontents*}
\csvautobooktabular{stats_beta4_2_0_2_2.txt}
\caption{Parameters and circuit bit size set by Concrete TFHE compiler for four different sequence lengths (T) processed by the two alternative Transformer attention mechanisms. Note the difference in bit-precision that is required (last two columns).}
\label{table:stats_encrypted}
\end{table*}

\paragraph{Benchmark comparisons.} 
We carried out numerical experiments that trained Transformer models based on the new Inhibitor and the conventional dot-product attention on four standard tasks. The aim was not to achieve SotA results but rather to examine if the Inhibitor mechanism would perform comparably on a set of familiar tasks, which is why we used simple set-ups without hyperparameter tuning. From Table \ref{table:results_summary}, which reports the results, we note that for each task, the two alternative attention mechanisms score very similarly. Indeed, none of the differences are significant at 95\% confidence (over at least 20 repetitions for each experiment). 
For experiments with the Inhibitor, we used a shifted score with $\alpha=0.5$ and scaling factor $\gamma=\sqrt{d}$.

The \textbf{adding} problem was introduced by \cite{Hochreiter_1997} to test long-term memory for RNNs. 
It takes two inputs: a sequence of random numbers and a two-hot sequence (which, in our experiments, each is 100 long). The ground truth is simply the dot-product of the two inputs as vectors, which, rather obviously, is not a challenging task for a conventional (dot-prod-based) Transformer. It is encouraging that the addition-based Inhibitor performs just as well on this task (which is hard for, e.g., simple RNNs).

We also train one-layer Transformers for the \textbf{MNIST} handwritten digit recognition task \cite{lecun-mnisthandwrittendigit-2010} and \textbf{IMDB} movie-review sentiment analysis task \cite{imdb}, which are simple go-to benchmark task for image classification and text analysis, respectively.
Although far from SotA, we note that our simple models achieve decent accuracy for both attention mechanisms, where the differences in performance are not significant.

The final task, labeled \textbf{IAMW} in Table \ref{table:results_summary}, uses the IAM Handwriting Database \cite{IAM_database_2002}, which is a collection of handwritten words written by more than 700 writers.  
The model first applies two convolutional layers followed by a Transformer and uses 
Connectionist Temporal Classification (CTC) loss \cite{Graves2006}
as an endpoint layer to predict the text.
Edit distance \cite{Navarro2001} was used as the evaluation metric. Again, the differences are small (and not significant at 95\% confidence).

\paragraph{Scaling experiments.}
Next, we wanted to examine and compare the scaling properties for the proposed mechanism under two identified scenarios: i) quantization with integer arithmetics and ii) homomorphic encryption. 
Therefore, the two alternative attention mechanisms were implemented directly in low-level code rather than high-level ML libraries, where built-in optimizations for conventional models and design choices would bias a comparison.

\begin{table}
\centering
\small
\begin{filecontents*}{timing_short.txt}
Timing Plaintext,32,64,128,256
Dot-prod Attention,98.6 µs,330 µs,1.2 ms,4.48 ms
Inhibitor Attention,63.1 µs,178 µs,577 µs,2.5 ms
\end{filecontents*}
\csvautobooktabular{timing_short.txt}
\caption{Estimated plaintext execution time on CPU for four different sequence lengths (with fixed size single head).}
\label{table:results_timing}
\end{table}

For the plaintext experiment, we used integer 16-bit arithmetics implemented in the Rust programming language, which gives detailed low-level control over circuits and supports advanced time benchmarking through the Criterion package.
The encrypted implementation uses the TFHE scheme \cite{Chillotti_2019} for Fully Homomorphic Encryption on the Torus as supported in the Concrete library for Python \cite{Concrete_Python_Compiler}.
TFHE permits the evaluation of arbitrary single-variate activation functions under homomorphic encryption through the Programmable Bootstrap \cite{PBS}. 
We here used considerably smaller networks where the embedding dimension was limited to size 2. The encrypted circuits were implemented for integers with up to 8-bit precision.

\begin{table}
\centering
\small
\begin{filecontents*}{timing_short_beta4_2_0_2_2.txt}
Timing Encrypted,2,4,8,16
Dot-prod Attention,2.68 s,22.4 s,107 s,828 s
Inhibitor Attention,0.749 s,8.56 s,23.8 s,127 s
\end{filecontents*}
\csvautobooktabular{timing_short_beta4_2_0_2_2.txt}
\caption{Estimated encrypted execution time on CPU for four different sequence lengths (with fixed size single head).}
\label{table:results_encrypted}
\end{table}

The results indicate that the proposed Inhibitor mechanism can have a significant advantage, with i) 30\%--50\% saving for the plaintext implementation on CPU as per Table \ref{table:results_timing}, and ii) a factor 3--6 under encryption with TFHE as per Table \ref{table:results_encrypted}.
Timing estimates are averaged over 20 repeated experiments and significant at the 95\% confidence level. 
Table \ref{table:stats_encrypted} displays compiler parameter settings for the differnt encrypted circuits. Note from the last two columns that the dot-prod based variant requires up to two bits higher precision than the Inhibitor.
It also requires about twice as many PBS, which further contributes to it slower performance.

\paragraph{Conclusions.} 
The proposed Inhibitor Transformer allows straightforward quantization by replacing conventional dot-product attention by a mechanism based on absolute value (eq.\ \ref{eq:manhattan}), subtraction and ReLU function (eq.\ \ref{eq:inhibition}) that all directly support an integer implementation. 
Our scaling experiments indicate that it has considerable potential for computational savings both for Transformers processing plaintext data as well as ciphertext.
The speedup is obtained by replacing the \textit{query-to-key} matrix multiplication and the subsequent Softmax with a light-weight Manhattan distance attention score and the inhibitor activation.

While experiment results are promising on simple training tasks, \textit{for future work}, it is necessary to examine performance under more challenging settings, for example, by pre-training a much larger Transformer model and testing on modern NLP and image recognition benchmark tasks, both in plaintext and under homomorphic encryption. 

\section*{Appendix}

\subsection{Signed Inhibitor}
The inhibition mechanism described in equation \ref{eq:inhibition} can only produce a non-negative attended value matrix $H$, while the conventional dot-product attention can select from $V$ that have both positive and negative entries. We can rewrite inhibition to handle signed values simply as
\begin{equation}
    H_{ik} 
    = \sum_j \left( V_{jk}^+ - Z_{ij}  \right)^+
    + \sum_j \left( V_{jk}^- + Z_{ij}  \right)^-,
    \label{eq:signed}
\end{equation}
such that the ReLU terms within the sum will attenuate or extinguish $V_{jk}$ for large $Z_{ij}$, while $V_{jk}$ will pass through unaltered for $Z_{ij}=0$. 

\iftrue
\subsection{Implemention}
A naive implementation of equation \ref{eq:manhattan} and \ref{eq:inhibition} relies on expanding the dimension of the broadcasted differences, before passing the result through an $\mathrm{abs}$ or $\mathrm{ReLU}$ function and then summing over an inner dimension. This uses large amounts of memory and should be avoided. 

The ML software that we used for this study supports fused operations for certain operations that avoid memory bloat and make the execution of the inhibition much more efficient on CPU and GPU hardware. Specifically, it supports calculating the pairwise distance between two tensors\footnote{The \code{cdist} function in the PyTorch library for deep learning.}, which directly implements calculating the Manhattan distance of equation \ref{eq:manhattan}. 

For the inhibition step of equation \ref{eq:inhibition} we use a simple trick based on 
\begin{equation}
    x^+ 
    = \frac{x + |x|}{2}
    \label{eq:relu_trick}
\end{equation}
and write
\begin{align}
    H_{ik}  
    &= \sum_j \left( V_{jk} - Z_{ij}  \right)^+ 
    \nonumber \\
    &= \frac{1}{2} \sum_j V_{jk} 
    - \frac{1}{2} \sum_j Z_{ij}
    + \frac{1}{2} \sum_j |V_{jk} - Z_{ij}|
    \label{eq:cdist_inhibition}
\end{align}
where the first two terms are simple sums that do not expand memory usage, and the last term can benefit from efficient fused pairwise Manhattan distance implementations.

Similarly, we rewrite for the signed inhibition of equation \ref{eq:signed} to have
\begin{align}
    H_{ik} 
    &= \sum_j \left( V_{jk}^+ - Z_{ij}  \right)^+
    + \sum_j \left( V_{jk}^- + Z_{ij}  \right)^-
    \nonumber \\
    &= \frac{1}{2} \sum_j V_{jk} 
    + \frac{1}{2} \sum_j |V_{jk}^+ - Z_{ij}|
    - \frac{1}{2} \sum_j |V_{jk}^- + Z_{ij}|,
    \label{eq:signed2}
\end{align}
where we have, in addition to equation \ref{eq:relu_trick}, used the relation 
\begin{equation}
    x^- 
    = \frac{x - |x|}{2}
    \label{eq:nelu_trick}
\end{equation}
for the negative ReLU function to leverage a fused implementation.
\fi

\iftrue

\subsection*{Relation to Prior Work}

This version extends the original AAAI2024 submission~\cite{brannvall2024inhibitor_studentabstract} by adding material focused on the implementation and evaluation of the Inhibitor attention mechanism under Fully Homomorphic Encryption (FHE), using the TFHE scheme on the Torus. In addition to the original benchmark results, this paper provides a more detailed cryptographic context, including parameter selection and performance under encryption. While the core methodology remains unchanged, the scope has been broadened to put more focus on privacy-preserving machine learning applications.
\fi

\bibliography{main}

\end{document}